# A Review of Sentiment Analysis Research in Arabic Language

Oumaima Oueslati*, Erik Cambria†,
Moez Ben HajHmida*, and Habib Ounelli*

**Abstract**

Sentiment analysis is a task of natural language processing which has recently attracted increasing attention. However, sentiment analysis research has mainly been carried out for the English language. Although Arabic is ramping up as one of the most used languages on the Internet, only a few studies have focused on Arabic sentiment analysis so far. In this paper, we carry out an in-depth qualitative study of the most important research works in this context by presenting limits and strengths of existing approaches. In particular, we survey both approaches that leverage machine translation or transfer learning to adapt English resources to Arabic and approaches that stem directly from the Arabic language.

## 1 Introduction

By dint of the massive production of user-generated content in social media, sentiment analysis, also known as opinion mining, has become one of the most dynamic research fields in natural language processing (NLP) [79]. It has become a crucial element for decision-makers and business leaders as well as for the public users to understand sentiment and opinions. Because users increasingly tend to consult each other before making purchasing choices, decision-makers and business leaders are now making important investments in measuring public opinion about their products and services. They invest in sentiment analysis not only to keep their clients satisfied but also to further improve new products and services and attract prospective customers. This fact motivates the NLP community to increase its efforts towards sentiment analysis research. In [67], Cambria et al. suggest that one can use sentiment analysis tools to extract useful information from unstructured data. Their study evaluated the impact of sentiment scores or polarity detection on decision making and on several areas such as politics [68], economy [121, 134], and tourism [101].

Recently, the Arab world has become the focus of many multinational analysts as it represents an important player in the global economy and international politics [75]. Mining opinions about issues, like oil and gas prices, market movements, and politics, has received extensive attention, particularly after the Arab spring [109]. Concurrently, thanks to the increasing availability of social media platforms, the amount of Arabic texts available on the Internet has seen a significant leap. According to the Internet World Stats,[1] Arabic is the fourth most used Internet language after English, Chinese and Spanish. Around 185 Million Arabic speaking people use the Internet, which corresponds to 4.8% of all users. The number of Arabic-speaking Internet users has grown 1616.4% in the last seventeen years. In spite of this, Arabic sentiment analysis (ASA) research has not witnessed significant developments yet [19], mostly due to the lack of sentiment resources in Arabic [204].

---

*University of Tunis El Manar, Tunisia
†Nanyang Technological University, Singapore

[1]Internet World Stats -www.internetworldstats.com/stats7.htm

Thus, many studies have tried to use machine translation on English sentiment resources. This bilingual approach, however, is merely inept due to Arabic linguistic features, which are essentially different from the English language in terms of structure and grammar [83]. Arabic studies must consider the various dialects and the free writing forms that make research in this field more complicated. Most Arabic interactions in social media are produced in local dialects. However, monolingual studies, which treat Arabic texts, commonly ignore dialects and Arabic texts containing Latin letters. Most of the available resources and tools consider only modern standard Arabic (MSA). Thus, their reproduction is not relevant as it would lead to lower accuracy in real applications. According to Kia et al. [79], the real value of any sentiment analysis technique for the research community corresponds to the results that can be reproduced with it.

Several studies have been conducted to survey ASA research. They mainly aim to give readers a comprehensive overview of the state of the art works related to ASA. Among the most recent and relevant surveys, we cite Al-Ayyoub et al. [13] and Badaro et al. [44]. Al-Ayyoub et al. put a great effort in categorizing the largest number of articles regarding sentiment analysis tasks and resources, compared to the surveys conducted earlier. They followed a narrative (enumerative) approach to build their survey. However, this survey lacked an in-depth discussion of some important insights and future directions of ASA. Badaro et al. proposed a comprehensive survey that covers recent advances in Arabic opinion mining systems, Arabic NLP software tools, Arabic sentiment resources, and Arabic classification models.

In this paper, we review efforts in ASA research. We aim to outline important gaps in the literature and suggest future directions. Unlike existing surveys, we cover recent literature about all elements involved in the ASA workflow. We define and follow a logic roadmap in order to ensure the coherence between the different sections of the paper. In the end of each section, we provide an insightful discussion and we investigate the open challenges related to each ASA aspect. Furthermore, our review contains many examples from Arabic social media to ground discussed challenges in real-world use cases. These examples form a broad background for understanding the nature and challenges of real ASA that may motivate more researchers to contribute to ASA research. Additionally, unlike existing surveys that only highlighted the difference between Arabic and English, we also discuss other languages that share similar challenges with Arabic.

The remainder of this survey is organized as follows. Section 2 explains problem definition and review methodology. According to Denecke in [80], the main problem of multilingual sentiment analysis is the lack of lexical resources. Hence, after discussing ASA challenges in Section 3, we review Arabic sentiment resources in Section 4. In Section 5, we investigate monolingual sentiment classification in terms of different pre-processing techniques, feature extraction, and classification algorithms. Afterwards, in Section 6, we explore bilingual sentiment classification, specifically the use of English resources and machine translation techniques to process Arabic text. Lastly, Section 7 provides an in-depth discussion and detailed recommendations for future work.

## 2 Problem definition and systematic review methodology

Sentiment analysis is a NLP research field that focuses on analyzing people's opinions, sentiments, attitudes, and emotions towards several entities, such as products, services, organizations, issues, events, and topics. Sentiment analysis is a branch of affective computing research that aims to mine opinions from text (but sometimes also images [172] and videos [72]). Sentiment analysis techniques can be broadly categorized into symbolic and sub-symbolic approaches: the former include the use of lexica and ontologies [82] to encode the polarity associated with words and multi-word expressions; the latter consist of supervised [71], semi-supervised [115] and unsupervised [63] machine learning techniques that perform sentiment classification based on word co-occurrence frequencies. Among these, the most popular are algorithms based on deep neural networks [132], belief networks [70], randomized networks [114], generative adversarial networks [133], and capsule networks [211]. There are also some hybrid frameworks that leverage both symbolic and sub-symbolic approaches [62, 137].

Sentiment analysis comprises a multi-step process namely data retrieval, data extraction and selection, data pre-processing, feature extraction, and sentiment classification. The ultimate sub-tasks of sentiment classification are polarity classification [191], intensity classification [11], and emotion identification [196]. Polarity classification aims to classify text as positive, negative, or neutral. Intensity classification seeks to identify the polarity degree (e.g., very positive, positive, fair, negative, very negative). Further, emotion identification attempts to identify the specific emotions behind sentiments such as sadness, anger, and fear. Despite the many different labels, these three sub-tasks share a similar workflow to achieve their purpose (Figure 1).

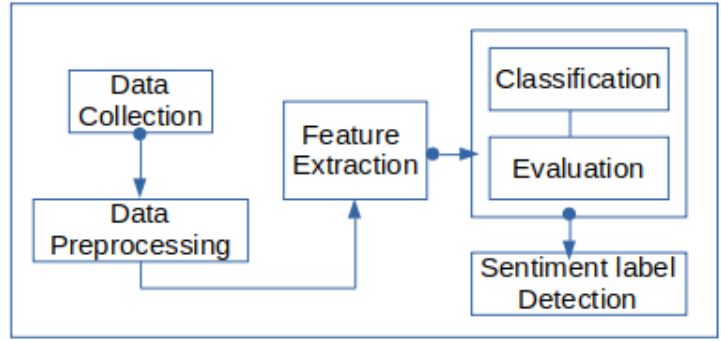

Figure 1: General workflow of sentiment analysis

Given the scarcity of Arabic sentiment resources [204], creating Arabic corpora and sentiment lexica is a major branch of related research. The resource quality profoundly influences the classification performance. For example, using a MSA corpus to classify dialectical texts may lead to poor accuracy. Likewise, pre-processing Arabic texts the same way as English texts may engender ambiguity. For example, stemming does not fit the Arabic language. The two terms 'لاعب' and 'يتلاعب', which in English mean 'to play' and 'to tamper', respectively, have the same basic stem in Arabic: 'لعب'. Hence, sentiment resources quality together with pre-processing techniques adaptation is critical for ASA [83]. Besides using sentiment resources and pre-processing techniques, ASA research also includes the use of machine learning. Machine learning methods treat sentiment classification as a multiclass or binary classification problem. While traditional machine learning techniques require hand-crafted features, recent algorithms can learn relevant features automatically.

In this survey, we review both monolingual and bilingual (translation from English) approaches and focus on the success and challenges of building sentiment analysis systems for the Arabic language. Our review follows the guidelines described by Kitchenham and Xiao in [131, 199] for conducting systematic reviews. Hereafter, we present the research questions (RQ) designed for our systematic review regarding the problem definition discussed above.

*RQ1:* **What are the major challenges in ASA?** This question is answered in Section 3. The motivation behind this research question is that several works take the bilingual approach. Discussing major challenges of ASA may incentivize future researches in this field to adopt language-specific techniques.

*RQ2:* **Do existing Arabic sentiment resources respond to major ASA challenges?** This question is answered in Section 4. The motivation behind RQ2 is that the Arabic space in social media is well-known for the use of informal writing such as Arabizi and local dialects rather than formal writing (MSA).

*RQ3:* **What are the approaches used to perform ASA in the monolingual setting?** This question is also answered and discussed in Section 5. The aim is to highlight works which initiate the work on informal Arabic content, and explore state of the art classification techniques.

*RQ4:* **How were English sentiment resources and techniques adopted to handle ASA?** This question is investigated in Section 6. The motivation behind RQ4 is to explore the weaknesses of bilingual approaches due to the differences between English and Arabic.

Reviewed papers in the different sections are obtained through querying several databases, such as Springer, Elsevier, IEEE, ACM, and ACL, using the following sets of keywords:
(Arabic + sentiment + analysis), (Arabic + dialects + sentiment),
(Arabizi + sentiment + analysis), and (Arabic + sentiment + resources).

Using these queries, Figure 2 and Figure 3 show the percentage of resulting papers per database and the number of reviewed studies for ASA relative to the year, respectively.

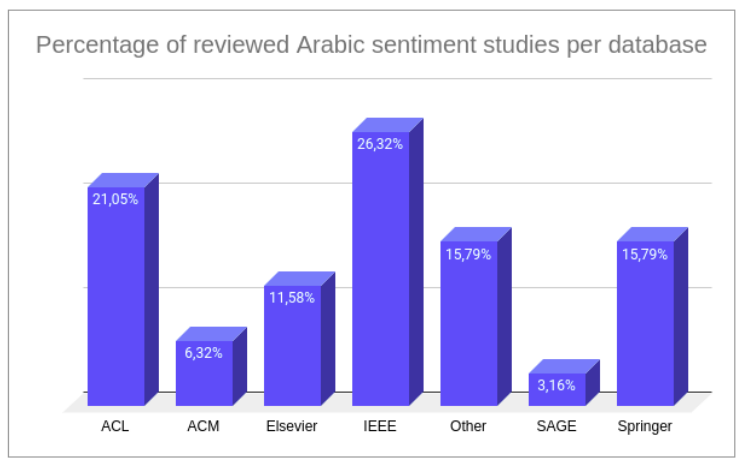

Figure 2: Resulting papers percentage per database

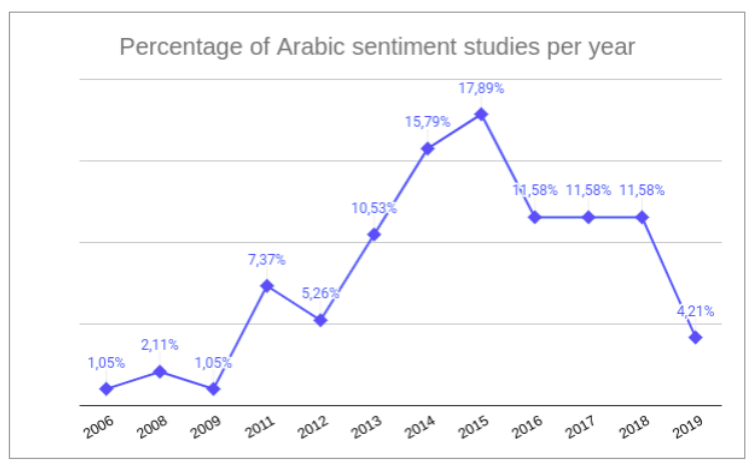

Figure 3: Resulting papers percentage per year

A recent bibliometrics survey of sentiment analysis literature, presented by Keramatfar and Amirkhani [124], reported and studied the distribution of 80 articles that investigated sentiment analysis in different languages. Using the above queries, we selected more than 100 papers on ASA in order to answer ASA aspects behind the designed research questions. Using similar queries for English rather than Arabic, we got around 12k papers. This confirms that ASA is still in its infancy compared to English sentiment analysis.

The selected papers for this review are dated from 2006 to 2019. Attempting to give a quantitatively overview of ASA literature since 2006 until 2019, we present in Figure 4 the studied ASA aspects by year. Upon a visual inspection, we note that modern standard Arabic is significantly more studied than dialects varieties and Arabizi. Although the Arabic space in social media tends to be informal using local dialects and Arabizi, dialects have started attracting interest since 2014.

However, Arabizi is still not fully investigated. On the other hand, the majority of the studies used machine learning techniques and algorithms to build ASA systems or to develop sentiment resources. Deep learning techniques have started to attract interest for ASA only since 2017. Through this review paper, we detail and discuss the different aspects of ASA in order to extract relevant research gaps and propose potential directions for future work.

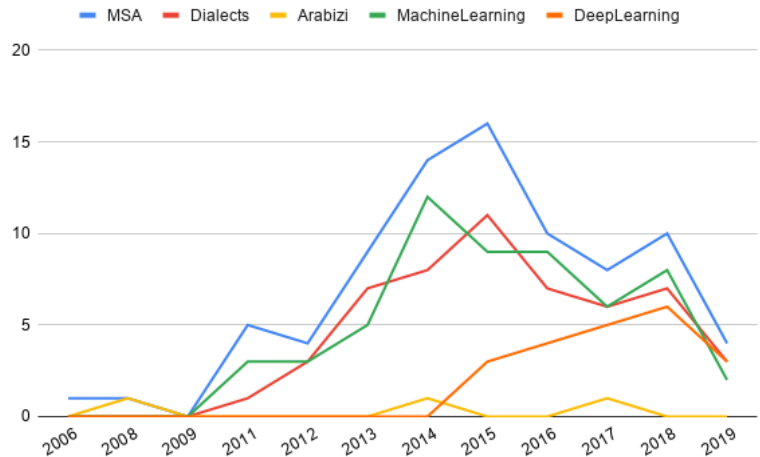

Figure 4: Arabic sentiment analysis research aspects from 2006 to 2019

## 3 Challenges in Arabic sentiment analysis

In this section, we present the challenges faced by ASA. Based on the literature, we first detail problems generally related to sentiment and subjectivity such as domain dependency, sarcasm, and spam. Then, we study challenges specific to the Arabic language. In Figure 5, we give a general overview of these kinds of challenges.

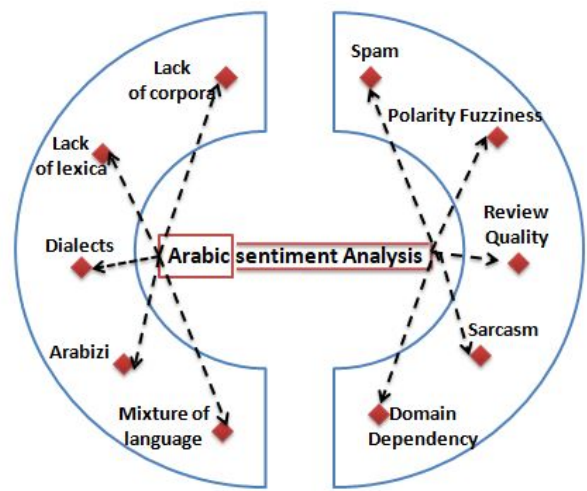

Figure 5: ASA challenges

## 3.1 General challenges

In [64], Cambria et al. argue that sentiment analysis is not a simple classification problem but a suitcase research problem that requires tackling many NLP tasks including subjectivity detection [69], aspect extraction [164], word polarity disambiguation [198], time expression recognition [212], and commonsense reasoning [61]. In [169], the authors detailed the challenges faced by classification techniques in sentiment analysis. In this section, we review the most common problems namely spam, polarity fuzziness, review quality, sarcasm, and domain dependency with examples.

***Spam:*** Through the growing popularity of online reviews in social media and their impact on reputation, illegal activity has become very developed. This activity consists of writing false reviews to promote or to damage the reputation of such service or product. It may also be a parasite advertisement which would benefit from the popularity of such webpage or Facebook page, etc. Figure 6 shows a sample of a spam review from a Facebook Page of a famous Arabic singer. While the singer post aims to get feedback about his new song, the review presents a job opportunity for stay-at-home women.

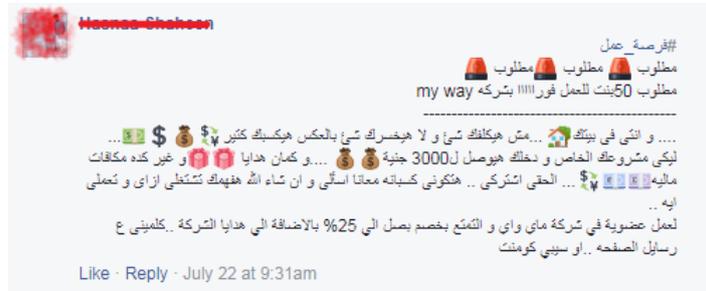

Figure 6: A sample of spam review from Facebook page

Spam detection for Arabic is still ongoing research: we only found a few works in the literature. In [193], the study was based on 3090 Arabic opinions collected from the social network Yahoo-Maktoob.[2] The classification was only based on URL presence: reviews containing an URL were classified as a spam. Evaluating this method with support vector machine (SVM) showed favorable results but using only URL filtering may not be efficient. In [107], the authors used 2848 Arabic reviews collected from online accommodation booking websites. They proposed a supervised approach for detecting opinion spam in Arabic. By combining techniques from data and text mining, their system showed high accuracy.

***Polarity fuzziness:*** Usually, opinion classification systems consider the polarity as positive, negative or neutral. It may happen that two humans do not annotate the review in the same way which makes polarity identification difficult. For example, while using the emoticon-based approach, a review may contain positive and negative emoticons at the same time. Also, the review can contain positive (or negative) emoticon and negative (or positive) keyword at the same time. Figure 7 illustrates a fuzzy review which contains positive and negative emoticons, but also positive words. Another source of fuzziness is the diversity of dialects. For example, 'نخلص

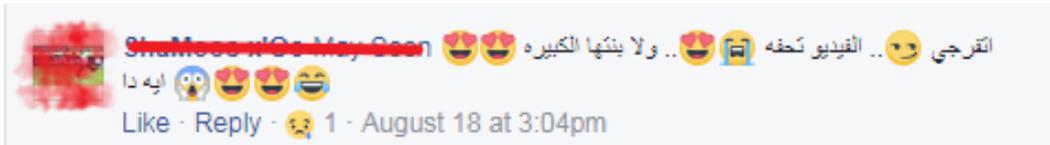

Figure 7: A sample of fuzzy review from an Arabic Facebook page

عليك' (nkhalass aalik), in Egyptian dialect it means 'I will kill you' which is negative. However,

---
[2]http://maktoob.yahoo.com

in the Tunisian dialect, the same sentence means 'I will pay for you', which is positive. To the best of our knowledge, there is no work until now which studies this problem for the Arabic language. However, there are few steps in this direction with efforts dedicated for identifying different dialects [57] and translating between them [110].

*Review quality:* Even a non-spam review may not be helpful for accurate polarity detection. The quality of the review is important to make sure the review is relevant to the opinion target. For example, comments like 'صباح الخير' (sabah el khir) which means 'good morning' may widely appear on social media pages. While positive, this kind of comment is not pertinent to evaluate the sentiment toward a product, a service or an organization. In [128], the authors proposed an algorithm to rank reviews by helpfulness, using a SVM regression system. A later work [170] tried to solve the problem by taking a concept-level approach, that is, by looking at both words and multiword expressions in the reviews. A very recent work [209] proposed a multilingual framework to predict review helpfulness. The framework was evaluated in the context of restaurants' online reviews, including variables generated by multilingual text processing programs. However, only English and German were tested with this framework. We could not find any study which considers review helpfulness in ASA.

*Sarcasm:* Sarcasm or irony is a form of speech that, in the context of sentiment analysis, mostly takes place when the speaker expresses a positive opinion but actually aims to complain about the opinion target [140]. Figure 8 illustrates an example of a sarcastic comment posted about an Arabic singer photo.

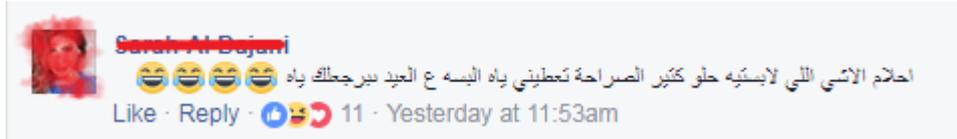

Figure 8: A sample of sarcastic review from an Arabic Facebook page

The review means 'you are wearing something very beautiful, lend it to me', but the user is actually making fun of the way the singer is dressed. According to [173], sarcasm is particularly hard to detect in Arabic language due to the use of positive indicators to express negative emotions. Hence, they recommended addressing this problem using contextual features. In [150], the authors reported that in their annotated MSA corpus, there are nearly 13.5% sarcastic tweets. However, the authors considered them as subjective tweets and they did not work on sarcastic features. Recently, the IDAT 2019[3] shared task was about detecting irony in Arabic Tweets [102]. Participant researchers had proposed several approaches, namely based on deep learning [123], ensemble learning [125, 154], and emotion classification [122]. Such a contest would motivate researchers to work on such remaining challenges, especially for languages with scarce resources like Arabic.

*Domain dependency:* Sentiment expressions vary among domains. Sentiment classifier trained to classify opinion in a domain may produce lower results when used in another domain [159]. The same sentiment word may have different polarity according to the domain. For example, 'كبيرة كثيراً' (very big) is positive if it is about accommodation. However, it is negative if it is about a dress. Many studies have tried to optimize general purpose lexicons for a particular domain using machine learning techniques [52, 156, 200]. For the Arabic language, building a general-purpose lexicon would lead to sub-optimal results due to the fuzziness coming from dialects. Focusing on two generalities at the same time (domain ad dialects) would be a more complicated task.

---

[3]http://www.irit.fr/IDAT2019/

## 3.2 Arabic-specific challenges

Besides the general challenges of sentiment analysis, there are other challenges related to Arabic varieties and morphology. As sentiment analysis depends significantly on the morphology of the target language [79], we list the linguistic properties of the Arabic language in terms of varieties, orthography, and morphology.

*Arabic varieties:* Arabic is one of the six official languages of the United Nations, and the mother tongue of about 300 million people in 22 different countries,[4] including standard Arabic and dialects. Arabic has three main varieties illustrated in Figure 9. The classical Arabic, also termed Quranic Arabic, is used in religious texts and many old Arabic manuscripts [184]. MSA is the formal language of communication understood by the majority of Arabic speaking people, as it is commonly used in radio, newspapers, and television. Dialectical or colloquial Arabic is used in daily conversation and recently used on both TV and radio.

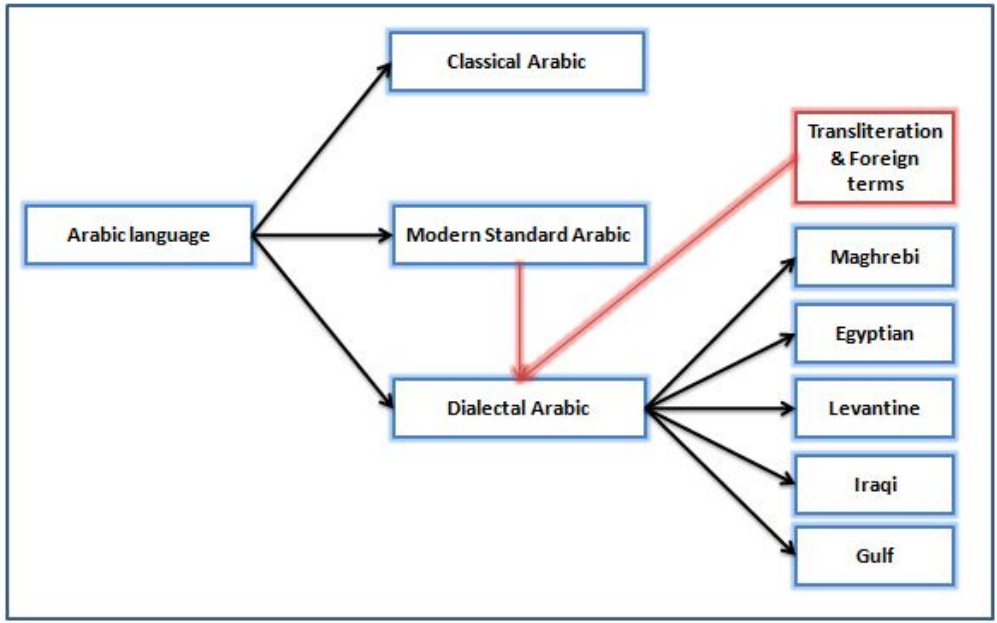

Figure 9: Arabic language varieties

There are mainly five groups of Arabic dialects: **Maghrebi**, which is spoken in North Africa including Tunisia, Algeria, Morocco, Libya, and Mauritania; **Egyptian**, which is spoken in Egypt and Sudan; **Levantine**, which is spoken primarily in the Levant region; **Iraqi**, which is spoken in Iraq, and **Gulf** which is spoken primarily in Saudi Arabia, UAE, Kuwait, Bahrain, Qatar, Oman and Yemen. This classification is general and based on the geographical location where nearest countries tend to have more common dialectical words and expressions. Dialects vary from one Arab country to another and from one region of the same country to another. For example, the Saudi dialect includes Najdi (Central) dialect, Hejazi (Western) dialect, and Southern dialect.

Most Arabic social media commentary is made using local dialects, as people prefer expressing themselves using their regional dialects [37]. Given that early technology did not fully support Arabic script, Arab people resorted to Arabizi [202], which is a form of transliteration that uses Latin characters to write Arabic texts. For example, the word 'newspaper', which is in Arabic 'جريدة', is written as it is pronounced 'jarida' in Arabizi. Moreover, the younger Arabic speaking generations are bilingual or even trilingual. They tend to use foreign terms so that another transliteration form has appeared. It consists of writing English or European terms using Arabic characters, especially with the evolution of technology, when writing Arabic had become technologically friendly.

---

[4]http://uicarmenia.org/en/2143

*Arabic orthography:* The Arabic alphabet includes 28 letters, as shown in Figure 10, with the absence of upper and lower cases. Arabic is read and written from right to left. It uses diacritical marks that can be placed above or under the letters to pronounce the word correctly and to clarify its meaning. However, proficient speakers do not need diacritical marks to understand a given text. Thus, the majority of MSA texts are written without these marks. As a result, a lexical ambiguity problem is created. For example, from the same undiacritized word 'قط', we can derive قطٌ, which means 'cat' and pronounced 'ketton', and قَطُ which means 'never' and pronounced 'kattoo'. This ambiguity is larger due to dialects varieties. For example, in Egyptian dialect, the word 'قُطْ' is pronounced 'ott' and means cat also.

| ا | ب | ت | ث | ج | ح | خ |
|---|---|---|---|---|---|---|
| alif | baa | taa | thaa | jiim | haa | kha |
| د | ذ | ر | ز | س | ش | ص |
| daal | thaal | raa | zaay | siin | shiin | saad |
| ض | ط | ظ | ع | غ | ف | ق |
| daad | taa | thaa | ayn | ghayn | faa | qaaf |
| ك | ل | م | ن | ه | و | ي |
| kaaf | laam | miim | nuun | ha | waaw | yaa |

Figure 10: Arabic alphabet

*Arabic morphology* The Arabic language is known by its morphological complexity and richness. The same word may carry important information using suffixes, affixes, and prefixes [186]. An Arabic word reveals several morphological aspects, namely derivation, inflection, and agglutination. The derivation consists of deriving from the same root several words with entirely different meanings. For example, from the root 'شعر' we can get 'شاعر', which means 'poet', and 'شعور', which means 'feeling'. The inflection aspect is the variation of the same word to describe the same meaning in several grammatical categories. For example, the verb to study which is in Arabic 'درس' (darassa) is 'أدرس' (adrosso) in the present tense with the first person singular and 'درست' (darassto) in the past tense. Lastly, agglutination means word attachment to affixes. For example, to express the English phrase 'for my sister' in Arabic, we need to attach to the noun sister 'أخت' the affix 'ي' which means 'my' and the affix 'ل' which means 'for'. Hence, the phrase is 'لأختي' and is pronounced (le-okhtee).

The complex morphology, as well as the language properties of Arabic, have created additional challenges for ASA, especially the lack of Arabic sentiment resources. Hereafter, we discuss the challenges related to the complexity and the diversity of Arabic language namely the lack of corpora and lexicon, the presence of dialects and Arabizi as well as the mixture of language.

*Lack of corpora:* A significant factor of an accurate sentiment analysis system is the use of large annotated corpora. The accuracy increases proportionally with the quality and the size of the used corpus to train the sentiment classifier. Arabic is poor in terms of corpora [204], and the scarcity of Arabic corpora for sentiment analysis is a well-known problem [19]. Additionally, the few available corpora are dialectical limited or even free from dialectical content. To the best of our knowledge, there are no Arabic corpora annotated for sentiment analysis and fully covering the different dialects. Even when considering one or a few dialects, the various writing forms are neglected. Having access to freely available corpora would improve ASA tools. In Section 5, we study corpora designed and annotated for ASA.

*Lack of sentiment lexica:* Many works have generated sentiment lexica from existing corpora [151–153]. The resulted lexica generally share the same limits as corpora, such as considering dialects and Arabizi. MSA lexica are small compared to English lexica. Accordingly, many works [1, 150] attempt to translate large English lexica to Arabic. However, the resulted coverage is poor regarding the morphological complexity of Arabic. Arabic sentiment lexica are further studied in Section 5.

***Use of dialectal Arabic:*** While people in social media express their opinions using their local dialects, the majority of NLP tools are designed to parse MSA [204]. Dealing with dialects make the task more complicated because there are no rules, no standard formats either. Because of colonization and other historical reasons, many dialectical words are derived from foreign languages such as French and English. Some other words are derived from standard Arabic but written differently using Latin letters. Table 1 illustrates the gap between MSA and Arabic dialects. The differences in words, syntax, and phonetics between MSA and dialects, and dialects themselves make the need to use dialect-based resources rather than using MSA text-based resources. However, existing resources are dialectical free, or very limited. One step towards bridging this gap is the use of machine translation between MSA and dialectal Arabic. Recent works [110] have shown promising results in this direction even if a parallel corpus does not exist [100]. Nonetheless, building customized dialectical parsers should be strongly considered in future works.

| MSA word | Dialectical word | Arabizi | Country | English equivalence | French equivalence |
|---|---|---|---|---|---|
| حلو / جميل jameel | حلو | 7elew | Lebanon | nice | beau |
|  | حلو | 7ilew | Saudi |  |  |
|  | حلو | 7low/ hlow | Tunisia |  |  |
| جدا jiddan | كتير | ktir | Lebanon | Very Wide | trés |
|  | وايد | wayed | Emirate |  |  |
|  | أوي | 2awi | Egypt |  |  |
|  | برشا | barcha | Tunisia |  |  |
| دراجة Darraja | بسكلات | Besklet | Tunisia | Bicycle | Bicyclette |
|  | دراقة | Darraga | Egypt |  |  |

Table 1: Dialectal Variation

***Code switching:*** Arabic users of social media tend to use Latin characters to represent Arabic words. This trend is known as Arabizi. The word Arabizi is originated from the portmanteau of Arabic and Englizi. Englizi is how the word English is pronounced in daily Arabic. This format is widespread in the Arab world, and most of the new generations use it for code switching, e.g., using Arabic and English or Arabic and French in the same review. Figure 11 shows a code switching example from Facebook that uses Arabizi and French in the same review. The review is "She is beautiful, may God bless her and bless you". The whole review is written using Latin letters. However, the first part 'she is beautiful' is in French language, and the second part is in Maghrebi dialect.

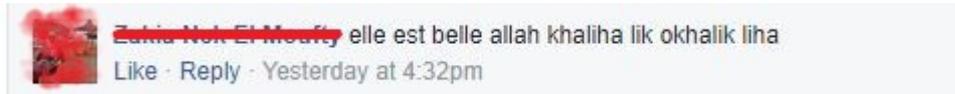

Figure 11: A sample of code switching (Arabizi and French) from an Arabic Facebook page

Given the above, filtering out Latin letters in order to conserve pure Arabic text would engender a loss of valuable information or the suppression of the whole review. Thus, dialect and Arabizi has the same degree of linguistic challenge in sentiment analysis: both have no standards rules and can be written differently.

***Code switching in different alphabets:*** Due to cultural facts, Arabic youth tend to be bilingual or multilingual. Therefore, one post can receive multilingual reviews that contain more than a language. Figure 12 shows a sample of a mixture of languages (Arabic and French). The review is "the best family" the word family is in French, and the word best is in Arabic. The review is not only a mixture of two different languages but also of two different alphabets: Arabic and Latin. In this case, pre-processing texts by filtering out Latin letters would cause meaning loss and weak classification.

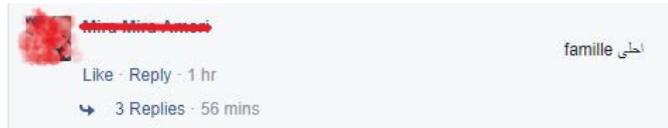

Figure 12: A sample of language mixing from an Arabic Facebook page

## 4 Arabic sentiment resources

### 4.1 Corpora foundations

A corpus is a body of texts, utterances, or other specimens. It is considered more or less representative of a language and is usually rendered in a machine-readable format [141]. Nowadays, computer corpora may store millions of running words with features that can be analyzed through tagging techniques. Tagging is the process of attaching, identifying, and classifying labels onto words and other formations. In sentiment analysis, the corpus is fundamental to train the sentiment classification system. It contains an enormous amount of emotional expressions in the form of words, phrases, sentences, and paragraphs. The formation of a corpus for opinion mining and sentiment analysis consists of three main steps: collection, annotation, and analysis.

Most of the corpora for sentiment analysis are collected from websites and social media platforms. These platforms provide insights into people's sentiments towards entities and their aspects, such as people, organizations, or objects. Widely used collection methodologies are Web crawling and scraping or calling of Web APIs, such as Google Reader's API, Twitter's API, and Facebook Graph API [56]. The annotation of the collected data requires a scheme definition in the perspective of data classification and analysis. For sentiment analysis, this step is challenging due to the lack of agreed model or theory. The typical approach is to annotate the sentiment polarity as positive or negative; and to annotate the emotions as anger, disgust, fear, sadness, surprise, joy, and love [167]. The annotation can be conducted manually through crowdsourcing [194], or automatically [58] based on lexica and emoticons, at the level of documents, paragraphs, and sentences. The annotation task is strongly affected by subjectivity. Therefore, it is recommended to use more than one annotator, and to measure and adapt the inter-annotator disagreement [56]. Cohen's kappa coefficient [42] is the commonly used measure to determine the disagreement. In [142], McCulloh et al. recommended the use of Krippendorff's alpha because it can be adjusted for a variable number of annotators assessing different reviews and handles missing data.

Annotated corpora for sentiment analysis are useful to train and test machine learning statistical tools for sentiment classification. The tools' performance strongly depends on both data quantity and quality. The reliability of the annotated data is usually evaluated by comparing its classification results with the human-annotation results. The analysis and exploitation of a corpus may reveal the limits of the annotation method or data sampling which can be respectively overcome by improving the annotation methodology or by collecting additional adequate data [56]. In [183], Sarmento et al. introduced a rule-based system to automatically build a specialized corpus for sentiment analysis. Experiments revealed that the negative comments are relatively easier to recognize, compared to the positive ones. This may be attributed to irony and polarity inversion. In 2011, Zhang et al. [208] built a corpus to deal with polarity shifting in English. In 2013, Bosco et al. [56] explored the central issues related to the development of an irony corpus for sentiment analysis in Italian.

For Arabic corpora, especially the dialectical ones, the analysis and exploitation bring additional complexity for the annotation step. Manual annotation is relatively expensive and time-consuming due to the presence of several dialects and foreign terms. Consequently, the corpus should be annotated by individuals from different Arabic countries to include all possible dialects. As an alternative, lexicon-based labeling can be used for annotation of Arabic corpora. However, it is still challenging due to the lack of Arabic sentiment lexica [1, 204]. In such a case, annotation based on positive and negative emoticons would be primarily useful [116, 175].

## 4.2 Designed corpora for Arabic Sentiment Analysis

A corpus is a primary key for accurate sentiment analysis since it is used to train and evaluate classifiers. In the last few years, many researchers made efforts to develop more available Arabic resources, such as corpora, which improve ASA development and evolution. However, it is still challenging to find Arabic corpora designed for subjectivity and sentiment analysis tasks. Table 2 shows the available corpora used in ASA. We selected only the most cited works and provided their resource URLs. We evaluate each corpus based on the following parameters: quantity, quality, coverage of dialects, Arabizi, attention to fuzziness, and availability. Resources are sorted by descending size. The quality column indicates if the authors used a specific approach to select only relevant reviews. Similarly, the fuzziness entry is checked if multiple annotators labeled the data.

| Corpus | Size | Dialects | Arabizi | Quality | Fuzziness | Public |
|---|---|---|---|---|---|---|
| BRAD [92] | 500k | ✔ | ∅ | ∅ | ∅ | ✔ |
| HARD [93] | 370k | ✔ | ∅ | ∅ | ∅ | ✔ |
| LABR [152] | 63k | ∅ | ∅ | ∅ | ∅ | ✔ |
| El Sahar et al. [96] | 33K | ✔ | ∅ | ∅ | ∅ | ✔ |
| Tunisian Corpus [143] | 17K | ✔ | ✔ | ∅ | ∅ | ✔ |
| AWATIF [2] | 11K | ∅ | ∅ | ∅ | ✔ | ∅ |
| ASTD [153] | 10K | ✔ | ∅ | ∅ | ✔ | ✔ |
| AraSenTi-Tweet [29] | 17K | ✔ | ∅ | ∅ | ✔ | ✔ |
| MASC [19] | 9K | ✔ | ∅ | ∅ | ∅ | ✔ |
| Refaee et al. [173] | 9K | ✔ | ∅ | ∅ | ∅ | ✔ |
| ArSentD-LEV [47] | 4K | ✔ | ∅ | ∅ | ✔ | ✔ |
| OCA [180] | 500 | ∅ | ∅ | ∅ | ∅ | ✔ |

Table 2: Arabic corpora designed for sentiment analysis

Opinion Corpus for Arabic (OCA)[5] is one of the earliest works on Arabic sentiment corpora construction. Rushdi-Saleh et al. [180] made efforts to build it by including a parallel English version called EVOCA manually. The corpus consists of 500 reviews, half negative and half positive. In the pre-processing, they filtered out unrelated comments, Arabizi, multilingual reviews, misspelled words, and out of scope reviews. The corpus is freely accessible for research purposes accompanied by unigram, bigram, and trigrams. Considering the limited resources for ASA at that time as well as the lack of Arabic language parsers, the efforts to construct OCA are pioneering and motivated researchers to build more improved corpora and prototypes. OCA limits are mainly the small size, the lack of any neutral or objective reviews, and the restriction regarding the covered domain.

Hotel Arabic reviews dataset (HARD) is the most recent corpus introduced by Elnagar et al. [93]. HARD is a rich dataset of more than 370,000 reviews expressed in MSA with some dialectical content. The balanced subset of HARD consists of 94,052 reviews made up of 46,968 positive and 47,084 negative reviews. Each review in the dataset is annotated as positive, negative, or neutral based on its rating. HARD is freely available to the research community in two forms: the unbalanced complete set and the balanced dataset. Although HARD is a large dataset, the presence of dialectical content is weak since the majority of the dialectal reviews are in Gulf dialects.

In 2012, Abdul-Mageed and Diab [2] released AWATIF, a multi-genre corpus of MSA labeled for subjectivity and sentiment analysis. The corpus consists of 10,729 sentences collected from different resources; 2,855 sentences from newswire stories, 5,342 sentences from Wikipedia talk pages, and 2,532 threaded conversations from web fora. For the annotation, the authors followed what they called simple guidelines as well as linguistically-motivated and genre-nuanced guidelines. They studied the effect of linguistic knowledge on annotation quality. Moreover, they considered the fuzziness in the task of subjectivity and sentiment. Hence, they aimed to train the annotators in the task adequately. However, their dataset targets only MSA which is not common when writing reviews on most websites and social media. Additionally, their resources are not publicly available.

---
[5]http://metashare.upf.edu/repository/browse/oca-corpus

More recent works have attempted to introduce the big data scale of ASA. As a first corpus, Aly and Atiya [152] presented a Large-scale Arabic Book Review (LABR)[6] with 63k records, only for a specific domain. Each record is a book review that had been written by a reader. LABR includes the users' ratings of books, scaled from 1 to 5. The authors considered the ratings of 1 and 2 as negative, ratings of 4 and 5 as positive, and rating of 3 as neutral. Their corpus lacks diversity regarding Arabic dialects and Arabizi. Although LABR contains large amount of reviews, it only covers MSA.

The authors in [25] extracted 1,513 reviews from LABR and conducted manual annotations to construct the Human Annotated Arabic Dataset (HAAD), an aspect-based sentiment analysis (ABSA) resource. HAAD sentiment labels are four: positive, negative, conflicting, and neutral. HAAD covers four tasks related to ABSA: aspect terms extraction, aspect term polarity identification, aspect category selection, and aspect category polarity identification. The corpus included 1,296 distinct aspect terms and is publicly available to serve as a benchmark for ABSA research in Arabic.

Motivated by LABR, Elnagar et al. [92] present the largest Book Reviews in Arabic Dataset (BRAD) for sentiment analysis and other applications. BRAD consists of almost 510K book reviews. However, the balanced dataset comprises only 150K reviews. The corpus includes some dialectal content, mostly from the Egyptian dialect content. In a recent work [94], the authors introduce the second version of BRAD. They added around 200k new reviews in order to produce a balanced subset of BRAD with more than 200K reviews.

LABR's authors also presented the Arabic Sentiment Tweets Dataset (ASTD) [153]. It consists of 10K Arabic tweets, manually classified into objective, subjective positive, subjective negative, and subjective mixed. The dataset and the used experiments are publicly available.[7] However, these resources are limited to the Egyptian dialect only.

In [173], the authors presented a gold-standard annotated corpus to support sensitivity and sentiment analysis of Arabic Twitter feeds. At that time, it was the first publicly released corpus for this task. Around 9k tweets were collected from Twitter API and manually annotated by two annotators. The annotation showed an inter-annotator agreement of 0.816. For the classification process, they utilized a variety of linguistically-motivated feature sets, namely morphological, syntactic, semantic, stylistic, and social signals. According to the provided information, the corpus is multi-dialectical. However, they did not mention which dialects it covers.

El-Sahar et al. [96] collected more than 33k reviews to build a multi-domain Arabic resource that includes some dialects, particularly Egyptian, Saudi Arabian, and Emirati. The authors provided 1.5K reviews for movies (MOV dataset), 15k reviews for hotels (HTL dataset), 4.5K reviews for restaurants (RES dataset), and 15k reviews for products (PROD dataset). In this work, the rating reflected the overall sentiment of the reviewer toward the entity. The rating was extracted and normalized into positive, negative, or mixed. The authors eliminated all redundant reviews to avoid any irrelevant or spam review. However, this technique is very simplistic and cannot eliminate both spams and irrelevant reviews because a redundant review does not necessarily represent a spam. This technique may engender a loss of important reviews which tend to be shared and repeated by many users. Redundant reviews, which receive different ratings, should be explored to define annotation rules adequately. Besides the large size and dialects presence in the corpus, the authors share the resources, the detailed experiments and the source code for research purposes publicly.[8]

---

[6] http://github.com/mohamedadaly/LABR
[7] http://github.com/mahmoudnabil/ASTD
[8] http://bit.ly/1wXue3C

In [19], the authors provided a Multi-domain Arabic Sentiment Corpus (MASC), a new Arabic-language reference corpus. They made it publicly available to the research community.[9] The corpus contains 8,860 reviews from multiple domains and several Arabic dialects. The reviews were manually collected from several sources such as the Jeeran website, the Qaym website, Google Play, Twitter, and Facebook. The corpus contains reviews from 15 different domains. However, taking into consideration the diversity of dialectal Arabic, the size of the corpus is relatively small to cover several dialects on 15 different domains.

In contrast to the previous corpora of multiple dialects, some corpora have been developed for targeting specific dialects. In [143], the authors presented the Tunisian Sentiment Analysis Corpus (TASC), which mainly focused on the Tunisian dialect. They collected 17k users' comments from the official Facebook pages of Tunisian radios and TV channels, namely Mosaique FM, JawhraFM, Shemes FM, HiwarElttounsi TV, and Nessma TV. The corpus was manually annotated to positive and negative polarities. TASC is restricted to the Tunisian dialect and did not exclude the Arabizi texts, in contrast to the majority of previous work. The resources are freely available for research purposes in the same area.[10] Similarly, in [29], the authors introduced AraSenTi-Tweet, a corpus for ASA of Saudi tweets. The size of the corpus has reached 17,573 tweets annotated as positive, negative, mixed, neutral, or indeterminate. The corpus was manually annotated by three annotators, and kappa statistics were calculated to ensure the reliability of the annotations. In [47], the authors presented ARSENTD-LEV, an Arabic sentiment dataset that is composed of Levantine tweets to cover spoken dialects in Jordan, Lebanon, Palestine and Syria. The corpus consists of 4K tweets. The annotation process was performed via crowdsourcing, using the CrowdFlower platform. The annotators were first asked to select the overall sentiment of the tweet (very negative, negative, neutral, positive and very positive). Then, the annotators had to select sentiment indicators to identify whether the sentiment is expressed explicitly or implicitly in the tweet, and to specify the discussed topic.

Although many sources for ASA have been developed, there are still many issues that limit their usage and applicability. First, most of the recent works in ASA have not released their resources yet. Second, they share similar weak points such as the dataset size, the reviews' quality, and fuzziness consideration. Third, the majority of corpora are domain-dependent for Newswire and/or Arabic MSA only, excluding dialects and Arabizi. Among the few sources that include dialects, they only focus on one or two dialects. Additionally, except for TASC [143], they also exclude the Arabizi text. Fourth, only a few resources considered the fuzziness problem by multiplying annotators or including additional features. These weak points degrade the accuracy of the sentiment classification. To overcome these problems, we need to build larger resources to cover the rich morphology and the diversity of dialects. More importantly, these resources should give attention to the quality and the capacity of handling spam and fuzziness. The diversity of words, syntax, and phonetics, between MSA and dialects as well as among dialects themselves, may lead to a high degree of fuzziness. Therefore, future work should focus on building dialect-based corpora instead of MSA-based corpora, especially for the sentiment analysis task since people tend to use their local dialects and to be very informal.

## 4.3 Lexicon construction techniques

The manual construction of sentiment lexica is a labor-intensive and time-consuming task. In this section, we study the literature of lexicon construction, especially automatic or semi-automatic adaptation. In [127], the authors survey sentiment lexica construction approaches in detail. There are four main techniques: (1) manual construction, (2) bootstrapping from a set of seed words, (3) adapting a lexicon from another domain using transfer learning, and (4) machine learning or probabilistic learning, based on human sentiment coding.

**Manual construction** of sentiment lexica consumes too much time. In order to accelerate it, researchers use **crowdsourcing** to engage annotators on the Internet, such as Amazon's Me-chanical Turk.[11] In [148], the authors addressed two main issues of this technique for sentiment coding. First, the design of the sentiment coding task should be simple and clear and should be performed without training. Hence, the task does not need specialized knowledge. Second, the quality control has to be measured to identify coders who are not doing the task appropriately. To make the task more fun and attractive, researchers explored gamification. For example, Thisone et al. designed a two-player game [151] where the first player is the suggester and the second is the guesser. Given a review, the first player has to assess the sentiment polarity and to select the word that best reflects the polarity. Given the selected word, the second player is asked to guess the polarity of the review based on the word. This technique seems to be smart because it accomplishes two tasks: constructing a sentiment corpus with document-level sentiment coding, and deriving a sentiment lexicon from it.

---

[9]`http://github.com/almoslmi/masc`
[10]`http://github.com/fbougares/TSAC`

**Bootstrapping** a sentiment lexicon from a set of seed words can be accomplished by using the semantic relations between the seed words and other words in a lexical resource (e.g., WordNet or a thesaurus) [120], or the associative [190] or syntactic [112] relations with other words in a corpus. Then, a network of word relationships can be used to propagate sentiment values from seed words through the network. As an example, in [104], the authors used a network of words based on associative relations in a corpus. They used a set of 45 seed words (15 positive, 15 negative, and 15 neutral), and their sentiment values were propagated through the network.

**Adapting** or mapping a sentiment lexicon from one domain to another domain aims to optimize a general-purpose sentiment lexicon for a particular domain to obtain optimal results. For example, in [46], the authors first applied an existing sentiment lexicon to a Twitter corpus to calculate the sentiment values of tweets. Then, they identified new words that were strongly associated with positive or negative tweets. For the new words, sentiment values were determined by calculating the number of positive tweets containing the word minus the number of negative tweets.

**Machine learning** and probabilistic learning models were used by some researchers to identify sentiment-bearing words. For example, some works [201, 203] utilized Latent Dirichlet Allocation [54] to model the relationships between words, aspects, sentiments, and documents.

## 4.4 Existing Arabic sentiment lexica

The availability of annotated sentiment lexica is essential for the progress of sentiment recognition systems. Generally, it is time-saving and beneficial to annotate corpora and to disclose the opinion polarity. In this section, we review available Arabic sentiment lexica. Table 3 provides the main features of publicly available lexica. The size feature denotes the number of lexicon entries in terms of lemmas, synsets, or words. The attributes of Dialects and Arabizi indicate whether the lexicon includes informal Arabic content, or not. Similarly, the attributes of Translation and Public indicate if the lexicon is constructed based on English resources, and if it is publicly available, respectively.

Alhazmi et al. [34] mapped SentiWordNet (SWN) to Arabic. SWN is a publicly-available English tool. It assigns positive or negative numerical sentiment values to synsets. Consequently, the Arabic SWN sentiment follows the same scoring methodology as SWN 3.0. However, the Arabic version has limited coverage (10k lemmas), and it is not publicly available.

In the same direction, Badaro et al. [45] constructed ArSenL,[12] a large-scale Arabic Sentiment Lexicon, using a combination of SWN, Arabic WordNet [53] and SAMA analyzer [105]. The authors followed two approaches to build two lexica. The first approach mapped the Arabic WordNet entries into English SWN, while the second approach finds the highest overlapping synsets between SAMA's English glosses and SWN. The union of the two resulted lexica outperformed the state of the art Arabic sentiment lexica. Although this lexicon was considered as the largest Arabic sentiment lexicon in that time, it has only MSA entries, with no dialect words or Arabizi. Additionally, it is not developed from a social media context. Therefore, it may not be suitable for social media text with many dialects.

---

[11] http://muturk.com
[12] http://oma-project.com/ArSenL/download_intro

| Lexicon | Size | Dialects | Arabizi | Evaluated | Translation | Public |
|---|---|---|---|---|---|---|
| Alhazmi et al. [34] | 10K lemmas | ∅ | ∅ | ∅ | ✓ | ∅ |
| ArSenL [45] | 157k synsets, 28K lemmas | ∅ | ∅ | ✓ | ✓ | ✓ |
| Abdulla et al. [5, 6] | Three lexica: 4K words, 9K words, and 8K words | ✓ | ∅ | ✓ | ✓ | ∅ |
| Mahyoub et al. [139] | 7K words | ∅ | ∅ | ✓ | ∅ | ∅ |
| SentiRDI [145] | 18K words | ∅ | ∅ | ✓ | ✓ | ∅ |
| Duwairi et al. [84] | 16K words | ✓ | ∅ | ✓ | ∅ | ∅ |
| Bayoudhi et al. [49] | 8K words | ∅ | ∅ | ✓ | ✓ | ∅ |
| Al-Ayyoub et al. [12] | 120K words | ∅ | ∅ | ✓ | ✓ | ∅ |
| BiSaL [20] | 1K words | ∅ | ∅ | ∅ | ✓ | ✓ |
| ElSahar et al. [96] | 2K words | ✓ | ∅ | ✓ | ∅ | ✓ |
| Ibrahim et al. [117, 118] | 5K words, 3K idioms | ✓ | ∅ | ✓ | ∅ | ∅ |
| Alhumoud et al. [35, 36] | 4K terms | ✓ | ∅ | ✓ | ✓ | ∅ |
| SLSA [98] | 35K terms | ∅ | ∅ | ✓ | ✓ | ✓ |
| Aldayel et al. [33] | 1.5K words | ✓ | ∅ | ✓ | ∅ | ∅ |
| Duwairi et al. [86] | 2K words | ∅ | ∅ | ✓ | ✓ | ∅ |
| Khasawneh et al. [126] | 452 words, 194 emoticons, and 179 special symbols | ✓ | ∅ | ✓ | ∅ | ∅ |

Table 3: Arabic sentiment lexica

Following the same approach of ArSenL, a Sentiment Lexicon for Standard Arabic (SLSA)[13] is a publicly-available lexicon, proposed by [98]. SLSA was constructed by linking the lexicon of an Arabic morphological free analyzer Aramorph with SWN. Both ArSenL and SLSA utilized SWN to obtain the scores of words. However, ArSenL is not publicly available since it used the commercial SAMA analyzer. The evaluation of SLSA and ArSenL demonstrated the superiority of SLSA. Nevertheless, SLSA excludes dialects and Arabizi, and it cannot analyze user-generated content in social media.

Al-Rowaily et al. [20] developed a bilingual sentiment lexicon called BiSaL. They constructed two lexica: SentiLEn for English and SentiLAr for Arabic. The authors extracted sentiment words from 2,000 message posts of Alokab Web forum, related to cyber threats, radicalism, and conflicts. Then, three Arabic language experts annotated the extracted terms' polarity. For each term, they assigned a positive score [0, 1] and a negative score [0, 1]. If a term is used in both positive and negative contexts, the summation of its positive and negative polarity scores will be 1. The construction methodology was reported and explained. However, the authors did not provide information about the covered Arabic varieties and did not validate their resource in a real application.

In [139], Mahyoub et al. developed Arabic SSL using only Arabic resources. Initially, they built the lexicon from a small seed list of positive and negative words. Then, they expanded it by exploiting synset relations in a semi-supervised learning manner. These relations included both semantic and lexical relations. Lastly, they evaluated the lexicon on a movie corpus and a book corpus. The lexicon achieved higher accuracy compared to early-reported ones. However, the lexicon is limited to about 23k lemmas, and it is not publicly available. Also, it excludes dialect words and Arabizi.

---

[13] http://volta.ldeo.columbia.edu/~rambow/slsa.html

While users tend to use common phrases and idioms to express their opinions, these phrases do not necessarily contain sentiment-bearing words. Therefore, if these words are analyzed separately by any sentiment analysis method, the sentiment clue may not be detected. Hence, in [117], Ibrahim et al. initiated the effort to construct an lexicon for idioms and proverbs in the Egyptian dialect. They collected more than 32K idioms and proverbs from Arabic websites. Then, they selected more than 3K common phrases and manually annotated them into positive and negative. Based on similarity measures, they developed a technique to detect idioms in texts and to compute their polarity. They applied this technique to tweets and reviews, and they achieved a 98% accuracy for sentiment classification.

Mobarz et al. [145] exploited RDI-ArabSemanticDB, the Arabic lexical semantics database [43], to construct an Arabic Sentiment Lexicon. They used RDI-ArabSemanticDB relations to expand a seed list of positive, negative, and neutral words. These relations include approximately 150K Arabic words, 18K semantic fields, and 20 semantic relations. In order to evaluate their resources, they applied different machine learning classifiers of Arabic sentiment using a translated version of the MPQA lexicon [39]. They argued that the translation of an English resource does not output accurate results.

In [5, 6], Abdullah et al. constructed three lexica through three different methods. For the first lexicon, they selected 300 seed words from SentiStrength. Then, they translated them into Arabic using the English-Arabic dictionary. For each word, they added its synonyms and assigned them the same polarity of that word. They also extended the lexicon with Arabic dialects such as Egyptian, Khaliji, and Levantine. For the second lexicon, they directly translated the SentiStrength lexicon using Google translate. A human effort was made to normalize the lexicon, remove duplicates, and correct the errors. The third lexicon was generated from a labeled and balanced corpus of positive and negative comments. They applied a frequencies-based scheme to generate two lists of positive and negative words. The authors also considered dialects and semantic relations. However, the three lexica are size-limited and do not include Arabizi. Many other works [83, 84, 126] have been proposed for constructing a lexicon. However, they have the same limitations. They include some dialects, but they are size-limited and do not include Arabizi.

In conclusion, previous works for building Arabic sentiment lexica fall into two categories: 1) mapping an English sentiment resource to Arabic, and 2) applying machine learning techniques on Arabic resources. Most of these lexica utilized machine translation on English resources due to the natural complex of the Arabic language and to reduce the manual efforts. This commonly used technique does not perform well for ASA since the resulting resources do not cover the morphological diversity of Arabic adequately. Moreover, the machine translation approach still require manual efforts [4]. Therefore, like the available Arabic corpora, Arabic sentiment lexica are dialect-limited or dialect-free. Additionally, the majority of lexica exclude Arabizi and cannot be efficiently applied to the social media context. Future effort should focus on building a dialectical version of Arabic WordNet and provide more dialectical resources.

# 5 Monolingual sentiment classification

Arabic text has its specific characteristics, such as richness of the morphology and dialects, that largely differ from other languages. Thus, it requires specific pre-processing techniques and features. For the monolingual approach, shown in Figure 13, there are two main approaches: corpus-based and lexicon-based. The combination of these two can be referred to as the hybrid approach.

The corpus-based method, also called machine learning method, typically trains sentiment classifiers using a sentiment corpus and machine learning algorithms. Several supervised learning algorithms have been used to classify the sentiment label into positive or negative. Among them, SVM, Artificial Neural Networks (ANN), Naïve Bayes (NB), Decision Tree (DT), Logistic Regression (LR), and Maximum Entropy (MaxEnt) [50] are among the most used algorithms. These algorithms require hand-crafted features, such as part-of-speech (POS) tags and social media-driven features. Recently, deep learning techniques such as deep neural network (DNN), recurrent neural network (RNN), and convolutional neural network (CNN) emancipate researchers from feature engineering [136].

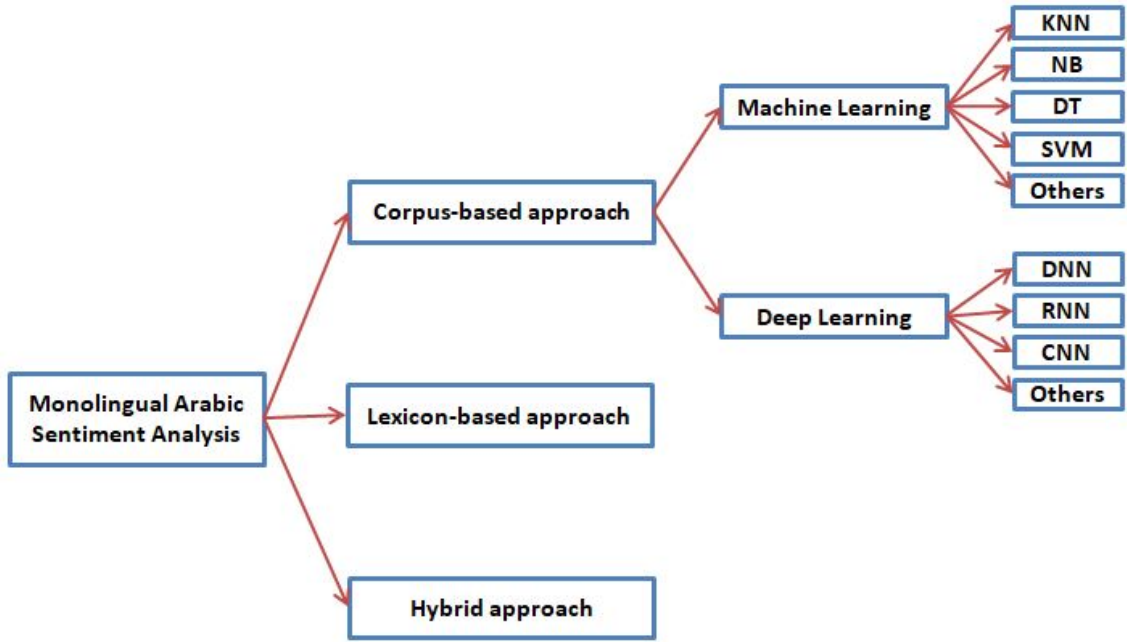

Figure 13: Monolingual sentiment analysis techniques

The lexicon-based method usually determines the sentiment or polarity of opinion by evaluating the sentiment words in the document or the sentence. It usually uses predefined dictionaries of annotated sentiment terms to label each word in the document by its sentiment. The hybrid or combination approach uses both lexicon and corpus-based methods. For that, the lexicon scores are used as input features to the classifier. Thus, sentiment lexica play an imperative role in the hybrid approach. The hybrid methodology is usually known to achieve a higher performance [48].

## 5.1 The corpus-based approach

The corpus-based method depends on labeled corpora and machine learning algorithms. The general methodology involves pre-processing the raw text, extracting the features, and training and testing the classification techniques. The whole process is illustrated in Figure 14.

### 5.1.1 Pre-processing

According to [40], the rich morphology of Arabic considers text pre-processing as an essential step for any Arabic NLP application. It includes filtering non-informative parts of the text. However, improper definition of what are these non-informative parts may lead the system to ignore important words [91]. ASA pre-processing is performed using tokenization [28, 83, 119], stopword removal [78], punctuation removal [90], upper/lower case conversion [210], word stemming [83], spelling check [78, 180], letter replacement (normalization), and dialect replacement. In Table 4, we present the widely-used pre-processing techniques for Arabic text and show how to apply them

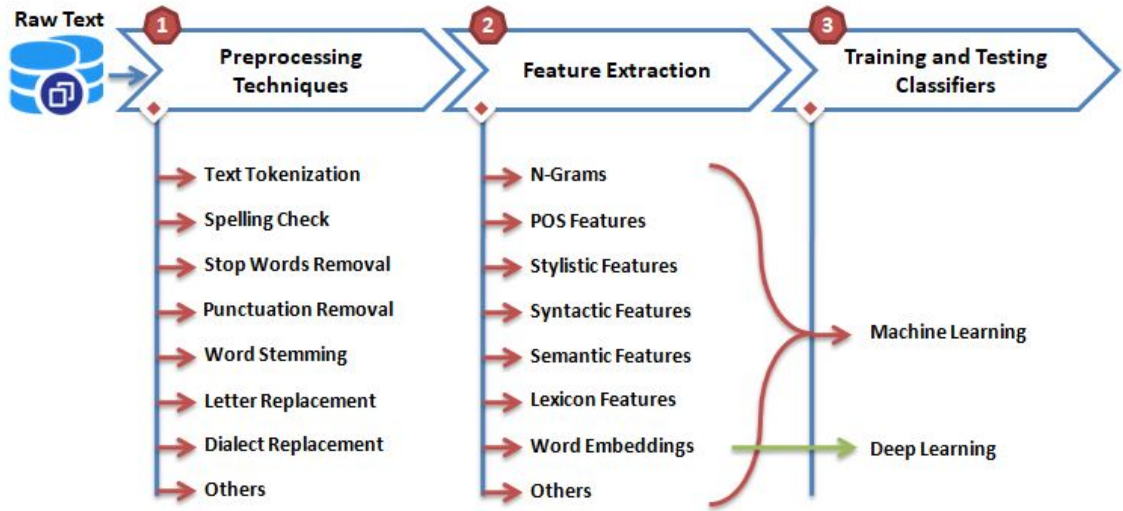

Figure 14: Machine learning-based methods for ASA

on a given example.

"أهلكت الأمطار و الرياح كللللللل المحلات."

| Pre-processing techniques | Description | Example |
|---|---|---|
| Text Tokenization | Tokenization is used to break a sentence into words, phrases, symbols, and characters. | أهلكت، الأمطار، و، الرياح، كللللللل، المحلات، . |
| Spelling check | Data coming from social media contain many spelling mistakes such as extra letters. | أهلكت، الأمطار، و، الرياح، كل، المحلات، . |
| Stop words removal | A stop word is an irrelevant term to the sentiment such as (كان،في) 'it was' and 'in'. Removal of stop words helps reducing the index space, and improves the response time as well as the effectiveness. In this case, we remove و which means 'and', and كل which means 'all'. | أهلكت، الأمطار، الرياح، المحلات، . |
| Punctuation removal | Most punctuation marks, such as full stops and commas, are not helpful to detect the polarity. | أهلكت الأمطار الرياح المحلات |
| Word stemming | Word stemming is about bringing each word into its basic form (root). Arabic stemming reduces the word to its three or four letters root. | هلك، مطر، ريح، محل |
| Letter replacement | In Arabic, some letters have different forms. So, some previous researches have replaced the different forms of each of these letters into the default form. | أ = ا، أ، ى |
| Dialect replacement | There are many studies which replaced dialect words with MSA words. | ∅ |

Table 4: Arabic text pre-processing

Several techniques have been introduced for pre-processing text. They follow different ap-proaches in order to preserve sentence semantics. For example, stemming methodology brings the word to its three-/four-letter root. However, it cannot always distinguish between root letters and affix letters [83]. Light stemming is another method that removes common suffixes and prefixes without reducing a word to its root. For Arabic texts, many researchers proposed to use light stemming since it avoids semantics loss, and it is faster than the root-based stemming approach. Stopword removal is a challenging step for text pre-processing due to the diversity of Arabic dialects. Darwish et al. [78] constructed a stopword list that includes 162 MSA and 90 dialectic words. However, it is still difficult to cover all the varieties where the same stopword can be written differently. Furthermore, social media data require additional pre-processing techniques since it contains emoticons, hashtags, and other chat-related text. The most common techniques are removing hashtags, removing URLs, identifying emoticons, and removing user mentions. Additionally, for Twitter, it is used to remove retweets and Twitter-specific characters.

### 5.1.2 Feature extraction

Feature extraction is a critical task in the domain of sentiment analysis. Selecting the right features determines the overall performance of sentiment classification [197]. The widely-used features for sentiment analysis are n-grams, POS tags, stylistic features, syntactic features, semantic features, lexicon features, and word vectors. In this subsection, we study how these features were utilized to perform ASA.

N-grams are sequences of n-items, extracted from texts. Generally, items are the words of the sentence. The most popular n-grams are unigrams, bigrams, and trigrams, which contain one word, two sequent words, and three sequent words, respectively. Many works have explored n-grams for ASA. In [158, 181], the authors showed that unigrams outperform both bigrams and trigrams. However, in [149, 180], the authors found that trigrams lead to the best performance.

POS tagging is the process of marking words, in a text, based on their nature and their relationships with adjacent and related words in that text. English POS tagging maps each word to POS of eight grammatical categories: verbs, nouns, pronouns, adverbs, adjectives, prepositions, conjunctions, and interjections. The English language has the luxury of Twitter-specific POS taggers, such as the CMU Twitter NLP tool [103]. However, most Arabic POS taggers are for MSA with some preliminary work for the Egyptian dialect [161]. Consequently, many works do not opt to use POS tags as features in their classification models [1, 173]. They reported that the currently-existing POS taggers are not suitable for social media text with enriched dialectal content.

Stylistic features aim to check the presence of specific sentiment indicators in the text, such as positive and negative emoticons, length distributions, vocabulary richness measures, and the frequencies of the special characters [173]. Stylistic features also include checking question marks and exclamation marks [130, 173]. Refaee and Reiser [173] explored the classified emoticons (positive and negative) in order to study sentiments in Arabic text.

Syntactic features are phrase patterns that indicate sentiments like nouns that are followed by a negative adjective. For example, the phrase "نهار اسود" (black day) means "a very bad day". This phrase is usually used in the Egyptian dialect. To extract such features, Refaee and Rieser [173] used n-grams of words and POS tags, lemmas, bag of words (BOW), and bag of lemmas as syntactic features. They reported that syntactic features depend on sentence construction and words assemblage. Alternatively, Al-Sabbagh and Girju [22] used transitive and intransitive verbs. Transitive verbs are defined as the ones that are found to be associated with object pronouns.

Semantic features represent the semantic orientation of the surrounding text. For sentiment analysis, given sentiment polarity, the semantic measures the correlation of a group of entities through different concepts of entities. For example, Saif et al. [181] used semantic features to map the entities into their groups. They showed that semantic features outperform the unigram and POS tagging features. Another example, Al-Sabbagh and Girju [22] used gender, number definiteness agreement between subject and verbs, nouns, and adjectives as semantic features. They argued that contrary to syntactic and stylistic features, semantic features involve human intervention.

Lexicon features are derived from sentiment lexica, such as polarity averages or sums [207]. While there are many general lexica for English, Arabic language counts few lexica that are automatically created from social media and translated from English lexica [182]. Many studies [130, 150, 173] on ASA reported that sentiment lexicon is an essential feature that enhances the classification performance.

Generally, the objective of sentiment classification is to divide text into positive and negative sentiments. The semantic and syntactic word distributions are known to be crucial challenges due to the hand-crafted features. In contrast, when the deep learning models are utilized for sentiment classification, word embeddings overcome this problem since they extract features automatically [41]. For the use of word embedding for ASA, there are two commonly used methods for word semantic distribution, which are word2vec [144] and GloVe [165]. Word2vec has been proposed for building word representations in vector space. It consists of two models: the continuous bag of word (CBOW) and skip-gram. GloVe is also used to build word representations, that trains its model based on the word-to-word co-occurrence from a corpus.

For ASA, only a few recent studies attempted evaluating word embeddings on Arabic texts. Zahran et al. [205] translated the English benchmark in [144] and used it to evaluate the different embedding techniques on a large Arabic corpus. However, they reported that translating an English benchmark is not the best strategy to evaluate Arabic embeddings. Dhaou et al. [77] built and evaluated Arabic word embeddings from a web-crawled corpus. They reported that the performance strongly depends on the quality of the data, and the high dimensionality vectors enhance the performance on a large corpus. Elrazzaz et al. [95] evaluated different embeddings methods and found that CBOW and GloVe consistently outperform other methods such as skip-grams. The most recent work of Alayba et al. [32] exploited the benefit of word embedding by applying word2vec in order to identify words similarity. Unfortunately, more recent word embeddings techniques, such as fastText [55], Embeddings from Language Models (ELMo) [166], Bidirectional Encoder Representations from Transformers (BERT) [81], are yet to be fully explored for ASA despite having pretrained Arabic versions of them publicly available, such as fastText for 157 languages [106],[14] Pre-trained ELMo Representations for Many Languages (ELMoForManyLangs) [74],[15] Arabic BERT,[16] Multilingual BERT (mBERT) [155],[17] etc.

### 5.1.3 Machine learning-based Sentiment classification

Many machine learning algorithms have been trained to conduct ASA. Their performance is generally measured through the prediction accuracy. In Table 5, we present the available works that utilize machine learning methods for ASA. For each work, we report its adopted pre-processing technique, extracted features, used classification algorithms, and performance.

---

[14] http://fasttext.cc/docs/en/pretrained-vectors.html
[15] http://github.com/HIT-SCIR/ELMoForManyLangs
[16] http://github.com/alisafaya/Arabic-BERT
[17] http://bertlang.unibocconi.it/

| Paper | Year | Language | Dataset | Pre-processing | Features | Accuracy |
|---|---|---|---|---|---|---|
| [186] | 2012 | MSA, Egyptian | Twitter | Removed noise, dialectal stopwords, and Arabizi | Unigrams and bigrams | SVM=72.1%, NB=65.4% |
| [18] | 2013 | MSA and dialects | Social media and news sites | Removed Arabizi and normalized Arabic alphabet | Domain features and sentiment features | KNN=90% |
| [83] | 2014 | MSA | News and Movie dataset | Stemming and light stemming | N-grams | NB=96.6% SVM=89.8% KNN=89.8% |
| [1] | 2014 | MSA and Dialects | Twitter, chat, forum | Tokenization and Lemmatization | POS Tags, standard, dialectal, and genre features | SVM= 85% |
| [7] | 2014 | MSA, Egyptian | Yahoo Maktoob website | Tokenization, stopwords removal, and stemming | Words frequencies (TF-IDF) | SVM=64.1% NB=55.6% |
| [118] | 2015 | MSA and Egyptian | Tweets and microblog | Removed Arabizi, symbols, numbers, stopwords | Standard and linguistic features | SVM=95% |
| [49] | 2015 | MSA | News and movie reviews | Word stemming, normalization and stopword removal | Opinion, stylistic, discourse, morphological, and domain features | SVM, MaxEnt, ANN = 85.6% |
| [153] | 2015 | MSA and Egyptian | Twitter | No pre-processing | n-grams | SVM=96.1% NB=68.2% LR=67.5% KNN=66.6% |
| [108] | 2016 | MSA | Hotel reviews, Social media | Word Stemming | PPS Tags | SVM, NB, DT = 96% |
| [21] | 2016 | Saudi dialect | Twitter | Remove tweets noise and tokenization | n-gram TF-IDF, and word occurrence | SVM=89.68%, NB= 82.7% |
| [85] | 2016 | MSA | Twitter and Facebook | Word stemming | n-grams, and TF-IDF | NB=69.78% SVM=66% KNN=50.98% |
| [17] | 2016 | MSA | Twitter | Spelling correction, proper nouns removal, and stemming | n-grams | NB = 64.85% |
| [38] | 2017 | MSA and Jordanian | Twitter | word stemming | TF-IDF and n-grams | SVM=88.7% NB=83.6% |
| [138] | 2018 | MSA and Moroccan | Facebook comments | Cleaning noise, normalizing, tokenizing, and word stemming. | TF-IDF and n-grams | NB=81.83% SVM=78.94% |

Table 5: Major works on Machine learning-based ASA

Although many machine learning classifiers have been used to conduct ASA, only three classifiers consistently showed superior performances: SVM, k-nearest neighbor (KNN), and NB. Duwairi and El-Orfali [83] studied these three classifiers using various pre-processing techniques and features. They found that the pre-processing step enhanced the classifiers' accuracy of Arabic sentiment, compared to the case of no pre-processing step. They also analyzed the performance of these three classifiers using word n-grams and characters n-grams. Results showed that n-grams in levels, words, and characters improved the performance. Duwairi and El-Orfali suggested that feature exploration with different classifiers is necessary for ASA.

While Duwairi and El-Orfali [83] and many other works [17, 49, 85, 108] targeted only the MSA, some works have been introduced for dialectical Arabic due to its widespread presence in social media. Al-Kabi et al. [18] developed a sentiment analysis tool for colloquial Arabic and MSA. They collected reviews from social media and news sites that include the Egyptian, Iraqi, Jordanian, Lebanese, Saudi, and Syrian dialects. For pre-processing, they removed Arabizi and normalized some Arabic letters. Then, they used KNN to classify the reviews, and they achieved an accuracy of 90%. To the same end, Abdul-Mageed et al. [1] developed SAMAR, a supervised system specialized for Arabic social media. They explored morphological features, standard features, dialectal Arabic features, and genre-specific features. Then, they classified the reviewers using SVM. They demonstrated that morphological information such as lexemes, lemmas, and POS tags has positive effects on Arabic sentiment and subjectivity analysis. Moreover, the standard features and the genre-specific features improved the performance of subjectivity and sentiment classification. On the other hand, the dialectal Arabic features decreased the performance of the classification models.

Many works [7, 118, 153, 186] have been proposed for both MSA and Egyptian dialects. Upon close investigation, we deduced that the accuracy strongly depends on the pre-processing techniques and the extracted features, in addition to the machine learning algorithm. Nabil et al. in [153] applied the SVM algorithm without any pre-processing technique, and they achieved an accuracy of 96.1%. In contrast, Ibrahim et al. [118] used the same SVM classifier after pre-possessing text by removing Latin letters. They achieved a lower accuracy of 95%, which may be attributed to the removal of Arabizi and losing essential sentiment words in the pre-processing step. Another work [7] used the SVM algorithm with simple features extraction. They achieved a low accuracy of 64.1%, which may be due to the simplistic definition of features based on frequencies.

The authors in [38] focused on MSA and Jordanian dialect. They applied supervised machine learning algorithms to Arabic user's social media of general subjects, which includes either MSA or Jordanian dialect. Using SVM and NB, they evaluated the impact of different weight schemes, stemming, and n-grams terms techniques and scenarios. Experiments showed that the SVM classifier using stemmer and Term Frequency-Inverse Document Frequency (TF-IDF) weighting scheme through bigrams is the best combination. This combination achieved the highest performance and outperformed the best scenario of the NB classifier.

A recent work by Maghfour et al. [138] analyzed the Facebook comments that are expressed in MSA and Moroccan dialectal Arabic. They aimed to examine the advantage of adding the classification of the Arabic corpus according to its forms (MSA or dialect), before the sentiment classification. Their main idea was to adapt text pre-processing based on each language category. For instance, the authors used light stemming for dialectical texts. They applied their approach with both NB and SVM classifiers. It enhanced their performance by considering the heterogeneity between MSA and the studied dialect. This two-step classification allowed to reduce errors caused by word stemming. However, this approach will be complicated in case of large and/or multi-dialectical datasets.

We note that the majority of machine learning-based approaches for ASA were mainly developed for MSA, and they used Twitter as a data source. Few works addressed the dialects, however, they were limited to one dialect such as Egyptian [7, 118, 153, 186] and Saudi dialect [21], except the work of Al-Kabi et al. [18] and Abdul-Mageed et al. [1]. Moreover, they all neglected Arabizi since they removed non-Arabic letters that include both the foreign terms as well as Arabizi.

### 5.1.4 Deep learning-based sentiment classification

Deep learning is a branch of machine learning. It is known for its ability to learn embedded and abstract representations from raw data with minimal human intervention. Deep learning has been recently considered for sentiment analysis and has shown excellent performance in this area [129, 185]. Deep learning models work efficiently and effectively with large datasets and save time because human intervention and feature engineering are not needed [31].

Deep learning models were successfully used to learn semantic representations of English text. These representations had led to accurate sentiment analysis. The most well-known deep learning models are CNN and long short-term memory (LSTM) networks. For ASA, only a few recent works have explored deep learning models. In Table 6, we present the significant works on deep learning-based ASA.

| Paper | Year | Language | Methods | Results |
|---|---|---|---|---|
| [24] | 2015 | MSA | DNN, DBN, DAE, combined DAE with DBN, and RAE | RAE= 74.3%, DNN=39.5%, DBN=41.3%, DAE with DBN=43.5% |
| [23] | 2017 | MSA and dialects | RAE, RNN | RNN=86.5% RAE=74.3% |
| [31] | 2017 | MSA and dialectical collected from Twitter | DNN and CNN | CNN=90%, DNN=85% |
| [15, 16] | 2017 | MSA and dialectical tweets | CNN, LSTM | LSTM=87.24%, CNN=85.01% |
| [157] | 2018 | MSA and dialects | Deep CNNs for processed and raw dataset, using character level features | 94.33% for processed dataset and 94.12% for raw dataset |
| [113] | 2018 | MSA and dialects | CNN, LSTM, and Ensemble Model (EM) based on the best the best CNN model and the best LSTM model. | CNN=64.30%, LSTM=64.75%, EM=65.05% |
| [14] | 2018 | Short dialectical text | Unidirectional and bidirectional LSTM and GRU. | LSTM=77.99%, GRU=78.71% |
| [76] | 2019 | MSA and dialects | combining differential evolution (DE) algorithm and CNN | DE-CNN= 87.64% |
| [26] | 2019 | Unspecified | LSTM networks | CNN = 82.7% |

Table 6: Major works on Deep learning-based ASA

Al Sallab et al. [24] investigated several Deep Learning models for opinion mining in Arabic that includes DNN, deep belief networks (DBN), deep auto-encoder (DAE), and recursive auto-encoder (RAE). The DNN model applies the back-propagation to a conventional neural network with several layers. The DBN layers are used as feature detectors. The DAE model reduces the dimensionality of original models. The RAE parses raw sentence words in the best order that minimizes the error of creating the same sentence words in the same order. Both DAE and RAE models provide a compact representation of the input sentence through unsupervised learning. Their objective is to minimize the reconstruction error of the input, without any manual annotations. The results showed that RAE outperformed all other models, followed by the DAE, featuring the advantage of recursive over one-shot models at learning accurate semantic representations. The reason is that RAE considers the context and order while parsing the sentence. On the other hand, DAE parses the whole sentence words at once in the first layer, with no consideration of the order of parsing words. However, RAE performance can still be enhanced by giving special attention to the linguistic characteristics of the Arabic language. Al Sallab et al. introduced a separate word embedding block trained on unlabeled instances from the QALB corpus. Using the DBN and DAE models, they reported the effect of sparsity on the learned representations and the overall performance.

Another study by Al Sallab et al. [23] addressed some limitations of RAE for ASA, such as its poor handling of Arabic morphological complexity. They introduced A Recursive deep learning model for Opinion Mining in Arabic (AROMA). AROMA starts by morphologically tokenizing the input, followed by the semantic and sentiment embedding models. The first stage of AROMA framework is unsupervised. It derived a vector representation for each sentence by applying an auto-encoder to its word sequence in a recursive manner. Then, they performed word sentiment embedding in order to provide a broader set of features that cover syntactic, semantic, and sentiment information. Then, they utilized the structure of automatically-generated syntactic parse trees to determine the order of the model's recursion. AROMA was evaluated on several datasets. The results showed that AROMA outperformed baseline RAE as well as other classic machine learning approaches, applied to ASA. The authors stated that the next version should explore the full orthographic and morphological space in Arabic, including different levels of surface representation and morphological abstraction. This aims to boost the performance of deep learning models for Arabic opinion mining and sentiment analysis.

In [31], Alayba et al. presented their Twitter dataset consisting of opinions on health services. The authors discussed the data collection and filtration process, pre-processing, and annotation processes. For classification, they conducted several experiments using various DNNs and several other machine learning algorithms (NB, LR, and SVM). The deep learning approaches showed promising results (90% for CNN and 85% for DNN) with the use of word2vec for the unsupervised learning. However, they were outperformed by other classifiers such as SVM, which achieved 90.88% of accuracy. Al-Azani et al. [16] focused on sentiment analysis of dialectical tweets from highly-imbalanced datasets. They extracted the main features by the word2vec model for word embedding. To address the imbalanced-datasets problem, they over-sampled the minority class by adding synthetic samples using the SMOTE [73] technique. In [15], the same group used CNN and LSTM networks with word embeddings features (CBOW and skip-gram). The proposed systems were evaluated on independent datasets, where the LSTM model achieved the highest accuracy of 87.27%.

In [157], Omara et al. made the first attempt to apply character level-based deep CNNs for ASA. This improved performance significantly compared to other machine learning classifiers namely KNN, SVM, DT, Random Forest (RF), NB, and LR. CNN achieved the highest accuracy of 94.33% while the accuracy of the second-best classifier was 87.09% as deep CNNs are powerful at extracting features from user-generated texts, either the pre-processed or raw ones. The authors in [113] explored different deep learning models that have not been applied to Arabic texts. They proposed an ensemble model that combines CNN and LSTM and they achieved significant accuracy.

In [14], the authors presented an empirical evaluation of deep RNN with non-verbal features. They proposed to detect sentiment polarity of Arabic microblogs based on emojis extracted features. They used the two forms of RNN: LSTM and gated recurrent unit (GRU) networks. For each network, they considered its unidirectional and bidirectional structures. Experimental results showed that LSTM and GRU based models outperform other classifiers such as SVM, KNN, and NB. Additionally, the bidirectional GRU achieved the highest accuracy with a slight difference from LSTM.

Dahou et al. in [76], proposed a framework, that adopts a differential evolution (DE) algorithm for CNN, in order to boost the performance. The DE algorithm was used to find the optimal configuration of both CNN architecture and network parameters. Five CNN parameters were tuned by the DE algorithm, including convolution filter sizes, number of filters per convolution filter size, number of neurons in the fully-connected layer, initialization mode, and dropout rate. The DE-CNN framework was evaluated on five Arabic sentiment datasets and achieved higher accuracy and lesser time-consuming compared to the state of the art algorithms.

A more recent work of Al-Smadi et al. [26] proposed an ABSA system that used a bidirectional LSTM (Bi-LSTM) and conditional random field (CRF). Integrating a Bi-LSTM with character-level embedding features has significantly enhanced the performance for aspect opinion target expression. The overall achieved accuracy to identify aspect sentiment polarity was 82.7%.

To summarize, semantic approaches and deep learning models are valuable tools for complex languages like Arabic. The multiple layers of deep learning models can effectively decompose the Arabic text, regardless of the complex rules of the language. However, while many works have applied deep learning models to sentiment text in English, only a few studies have been performed for ASA. Therefore, there is an urgent need for further exploring word embedding and deep learning models on Arabic sentiment, including also Arabic dialects and Arabizi. In fact, deep learning relies on the discovery that unsupervised learning provides to generate word representation. As word representations are the widely-used features for DNN, their quality directly impacts the DNN performance. Therefore, available unlabeled Arabic datasets should be explored further to ensure a better quality of these features. For example, exploring dialectical or Arabizi corpus to generate word vectors would allow an efficient ASA especially in the context of social media.

On the other hand, recent works such as [26] gave attention to ABSA for the Arabic language using deep learning. ABSA is concerned with review aspects and their corresponding sentiments, so that part of the text would be linked correspondingly to their aspects. Unlike document-level and sentence-level sentiment analysis, ABSA would not fail in cases such as "The dress quality is very beautiful, but the price is very expensive." In such a case, positive and negative polarities will be assigned to the dress quality and the dress price, respectively. The challenge of Arabic ABSA is mainly the lack of available tools to process and analyze Arabic content. ABSA was further experimented during the contests of SemEval-2016 Task 5 [59, 168, 178, 189]. Experimentation was conducted with English data (reviews) and other language including Arabic. The outcome of these contests is interesting as these provided new perspectives and resources for ASA research. However, Arabic is still left out from many interesting NLP tasks and the Arabic NLP community is working to bridge this gap. For example, stance detection in tweets was conducted during SemEval-2016 Task 6 where only an English Twitter dataset was proposed [147, 195, 206]. However, in the 2017 version [176], Arabic was added.

## 5.2 The lexicon-based approach

The lexicon-based approach is usually used when the data are unlabeled. Sentiment lexica are used to label the data and to predict the polarity. The use of sentiment lexicon allows assessing sentiment score for a document (a review) from the sentiment score of words or phrases in the lexicon [19]. However, as will be discussed in Section 6, only a few sentiment lexica are available for Arabic texts, and most of these lexica are not publicly available. According to Ortigosa et al. [160], the lexicon-based method is usually known as a weak method compared with machine learning methods. Therefore, only a few works used the lexicon-based method alone for sentiment analysis in the Arabic language. In contrast, most works have adopted the lexicon-based approach along with machine learning by designing lexicon features.

Al-Ayyoub et al. [12] presented a study that solely used the lexicon-based approach. They created a lexicon with 120K Arabic terms by extended the available lexicon in [9]. To evaluate the quality of their lexicon, they used a dataset of 300 tweets for each positive, negative, and neutral class. Then, they removed the repetition of vowels, fixed spelling mistakes, and fixed mistakes caused by sound similarities. Based on their lexicon, they achieved an accuracy of 87%. An older work of Al-Subaihin et al. [27] proposed a sentiment analyzer which classified reviews according to their sentiments indicators using sequence patterns and lexica. They tagged each word to POS, NEG, ENT, or NO, if the word is positive, negative, entity, or a negation, respectively.

## 5.3 The hybrid approach

The hybrid sentiment analysis approach combines both corpus-based and lexicon-based approaches. Only few works utilize the hybrid approach for ASA. Most of them employed lexicon-based tech-niques to label text polarity to be further used in the sentiment classifier. An early work of El-Halees et al. [89] presented a combined approach for ASA. The approach uses a lexicon-based technique to assign polarities to the collected unlabeled documents. The resulting labeled documents are used with a MaxEnt classifier as a training set. The aim is to classify the remaining documents that were not classified by the lexicon-based technique. Lastly, all the classified documents are fed into a KNN classifier to classify ignored documents by the lexicon-based and MaxEnt techniques. The accuracy reported by this approach was 80.29%.

Another work by Aldayel et al. [33] used a hybrid approach to classify Arabic tweets as positive, negative, or neutral. The used approach starts with dataset collection and data pre-processing. The pre-processing includes normalization, stopword removal, speech effect elimination, and stemming. Then, using a lexicon-based technique, pre-processed tweets are classified and fed to SVM classifier. This approach produced an accuracy of 84.01%. Likewise, Khasawneh et al. [126] used a hybrid approach to classify Arabic tweets as positive, negative, or neutral. The authors collected 1,500 Arabic tweets in three domains (economics, sport, and news). They labeled tweets using 13 lexica built manually. Using two machine learning classifiers, bagging, and boosting, the authors report an accuracy of 85.95%.

Some researchers such as El-Halees et al. [88] aimed to compare lexicon-based and hybrid approaches to analyze Arabic sentiment. The lexicon-based approach combined lexicon and MaxEnt and KNN machine learning algorithm. This combination has been reported as a more accurate approach than corpus and lexicon-based approaches [118, 188]. In [51], Biltawi et al. proposed a hybrid approach by presenting the review for the corpus-based approach in the same way it is seen in a lexicon-based approach. The main idea was to replace the words with their corresponding positive and negative labels in the lexicon. Thus, terms that are important but rare can be taken into consideration by the classifier. The authors conducted a deep comprehensive comparison using different classifiers and parameters. Results showed that the proposed hybrid approach outperforms the corpus-based approach. The highest accuracy was 96.34% using a RF classifier with 6-fold cross-validation.

Al-Twairesh et al. [30] proposed a hybrid method for sentiment analysis of a specific Arabic dialect, which is the Saudi dialect. They used a set of features that have been engineered to be dialect-independent and evaluated using a feature-backward selection method. Then, three classification models for Saudi Twitter sentiment analysis was developed and compared as follows: two ways (positive and negative), three ways (positive, negative and neutral), and four ways (positive, negative, neutral, and mixed). The authors investigated the impact of the proposed feature sets on all the developed classification models. They found that features extracted from the AraSenTi lexicon were present in all the best feature sets of all the experimented classification ways.

Recently, Elshakankery et al. [97] presented HILATSA, a hybrid Incremental learning approach for ASA. The main idea is to introduce a sentiment analysis tool for Arabic tweets able to cope with the rapid change of words and their usages. The authors built some essential lexica, namely words lexicon, idioms lexicon, emoticon lexicon, and special intensified words lexicon. To cope with different word forms and misspelling, Elshakankery et al. investigated the effectiveness of using Levenshtein distance algorithm. For the classification, two machine learning algorithms (SVM and LR) and one deep learning model (RNN) were used. Experiments showed promising results by showing high accuracy and reliable performance in a dynamic environment.

Although the hybrid approach has been reported as a more accurate approach than corpus and lexicon-based approaches, it is not yet explored for aspect-based ASA. Also, we note that deep learning models still not explored for hybrid approach. We find only the work of Elshakankery et al. [97] using RNN, however, the results were not largely discussed. Table 7 illustrates a summary of major works on the hybrid approach.

| Paper | Year | Language | Dataset | Pre-processing | Features | Accuracy |
|---|---|---|---|---|---|---|
| [88] | 2012 | MSA | Reviews about education, politics, and sports | Removed HTML tags, repeated letters and stopwords, normalizing Arabic letters, tokenization, and Arabic light stemmer | ∅ | MaxEnt, KNN = 80% |
| [90] | 2014 | Dialects | Twitter | Removed punctuation, numbers, special characters, and repetitions. Letter replacement. | Stems, Twitter, language-independent, and semantic features. | SVM= 84% |
| [188] | 2014 | MSA and Dialects | News websites and Facebook Pages. | Removed stopwords, stemming and data autocorrection | ∅ | SVM= 87% |
| [84] | 2015 | Dialects | Twitter | Tokenization, removed stopwords, and converting emoticons to their corresponding words | ∅ | With dialect lexicon: SVM=87% and NB=88% |
| [118] | 2015 | MSA and Dialects | Twitter and microblogs | Replaced idioms and proverbs with text masks | Standard, linguistic, and syntactic features | SVM = 95% |
| [187] | 2015 | Egyptian dialect | Twitter | ∅ | unigram, bigram and tri-gram | ∅ |
| [51] | 2017 | MSA and Dialects | OCA and Twitter | Noise removal, Arabizi removal, and normalization | PoS tagging and lexical features ( polarity and negation) | SVM, MaxEnt, BAGGING, BOOSTING, ANN, DT, NB, and RF=96.34%. |
| [30] | 2018 | MSA and Saudi dialect | Twitter | unspecified | Semantic, stylistic, and Twitter specific features. | SVM = 69.8% |
| [97] | 2019 | MSA and dialects | Twitter | Remove tweets noise, tokenization, removing Arabizi, and normalization. | Features from words, idioms , emoticon, and special intensified words lexicon. | LR=83.73%, SVM=81.52%, RNN=81.62% |

Table 7: Major works on hybrid ASA

Within SemEval-2016 Task 7, Refaee et al. [174] realized the top performing system for sentiment intensity for Arabic tweets. They used a hybrid approach based on training linear regression models on lexicon containing word-lemma unigrams. Then, sentiment intensity scores were adjusted based on a set of rules referring to existing sentiment lexica. The authors reported that high quality sentiment lexica are needed to accurately perform this task.

# 6 Bilingual sentiment classification

Most successful sentiment analysis systems for English rely on sentiment lexica such as SWN and SenticNet [65]. Thus, there have been efforts to take advantage of these lexica for ASA. This bilingual approach has mainly addressed mapping sentiment resources from English to Arabic, or translating Arabic text to English in order to assign sentiment polarity using English sentiment resources. Figure 15 highlights the bilingual techniques found in the literature.

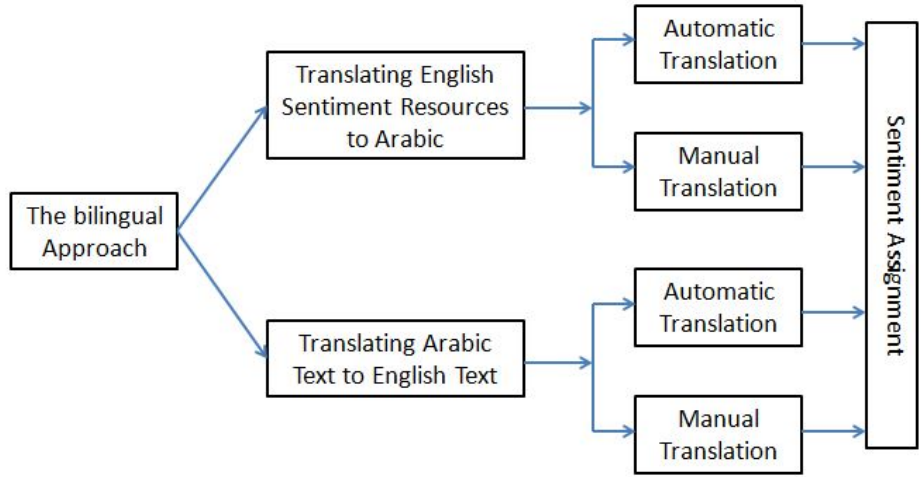

Figure 15: Bilingual sentiment analysis techniques

The first technique consists of translating sentiment annotated corpora and lexica from English into Arabic, and use them to assign sentiment polarity to Arabic text. The translation can be performed manually or automatically through translation APIs (Google Translation) [5]. For the manual techniques, the lexicon is built by translating the content of SWN [45, 98, 99, 179], SentiStrength [139, 171], or SenticNet [10, 111], and, for each translated word, finding its synonyms. Then, the word with its synonyms are added to the lexicon. For the automatic technique, the lexicon is built by starting from manually collected and annotated lexicon, called base lexicon. Then, the lexicon size is increased by adding synonym and antonyms. The manual method for lexicon creation is more accurate than the automatic one. However, the automatic method required less time and work [5].

The second technique consists on translating Arabic text into English, and using English sentiment resources to assign sentiment polarity to the text. For example, while the authors in [10] translated each entry found in the English version of SenticNet, the authors in [111] translated Arabic text, normalized and structured it, and then attempted to find a match in SenticNet. Both techniques, translating Arabic text and translating English sentiment resources share same weaknesses. Firstly, Arabic words that do not exist in the translated resources are ignored. Secondly, the quality of sentiment classification depends on the quality of the translation. Translation itself is challenging for Arabic language, especially for Arabic dialects. Last but not least, translation ignores dialectical texts or distorts its meaning.

# 7 Results and Discussion

The growing interest in ASA is due to the fact that the Arab world is a key player in international economics and politics, which is further amplified by the Arab spring. Since 2011, the Arab world has been the focus of many international analysts who track daily sentiments on issues like oil prices, terrorism, politics and election, marketing, and tourism. Therefore, there is a great need to tackle ASA challenges in order to realize qualitative sentiment resources and effective systems. In fact, besides the challenges of sentiment analysis, the main challenges related to ASA are the lack of sentiment resources and the varieties of Arabic dialects [96, 204]. The availability of sentiment resources such as lexicon and corpora is crucial to enable ASA progress. In the context of social media, ASA resources and systems challenge the growth of texts written in local Arabic dialects [19].

Earlier works on ASA are restricted to MSA only and do not tackle local dialects. MSA is morphologically rich and complex compared to English, but easier to parse compared to Arabic dialects. Unlike MSA, dialects do not follow standard rules and have a multilingual aspect [2, 3]. As MSA is mostly used on News, earlier ASA systems are based on size-limited resources. Earlier ASA systems failed in the context of social media, where dialects are extensively used [33, 126]. Some works had taken advantage of English sentiment resources prosperity, attempting to improve sentiment resources for the Arabic language. These works used a bilingual method based on English to Arabic translation and mapping techniques. The bilingual method had merely improved the size of Arabic resources. In the literature, many researchers criticized the bilingual method for producing sense ambiguity and being unable to deal with Arabic dialects.

While the bilingual method did not significantly improve the state of the art of ASA, researchers had to take other directions, mainly building dialectical resources. The aim is to allow existing ASA systems to deal with social media context. In the last three years, many researchers focused on building dialectical sentiment resources [96, 143]. Most of these efforts followed a three-phase workflow. The first phase consists of extracting dialectical texts from specific websites and social media platforms. The second phase applies various pre-processing techniques [40, 78]. Lastly, the third stage consists in assigning, manually or automatically, sentiment polarity to the extracted texts. Such a workflow produces resources that their quality depends directly on the applied pre-processing techniques. As researchers used to delete Arabizi texts during the pre-processing step [18, 23, 186], they missed valuable expressions and sentiment indicators. Exceedingly applying pre-processing techniques causes the loss of essential elements for understanding the specific meaning of the original text. We observe that most of the resulting resources focused on the Egyptian dialect [7, 186, 187], around 80% of resources. The reason behind this magnitude is the dominance of the Egyptian dialect on social media. In the case of a multidialectal context, using a mono-dialectical resource leads to weak classification performances. The meaning of a word or an expression depends on the specific dialect.

To leverage the complex morphology of the Arabic language and the diversity of dialects, we propose to shift from simplistic word-level sentiment analysis to concept-based sentiment analysis. Concept-based sentiment analysis is stepping away from methods based on keywords and word co-occurrence frequencies [60]. It takes into account the meanings of multi-word expressions based on the semantic analysis of the textual contents. For these methods, commonsense knowledge bases are the key to properly deconstruct natural language text into sentiments and to preserve the meaning carried by multi-word expressions [66]. As Arabic dialects are well-known for frequent use of multi-word expressions, future works should focus on building concept-based techniques and developing concept-based approaches.

There are currently numerous English concept-based sentiment resources. The most cited ones are SWN and SenticNet. SWN and SenticNet provide polarity values for each entry in the lexicon, namely sentiment orientation and degree. Recently, SenticNet has also focused on multilingual sentiment analysis research [135]. SenticNet was made available in 40 languages [192]. In particular, a lot effort has been devoted to Chinese sentiment analysis [162, 163]. Same as Arabic, Chinese is a complex and challenging language for the lack of resources. Arabic and Chinese have many common properties such as richness, varieties, and resource scarcity. Since SenticNet framework improved the state of the art of Chinese sentiment analysis, we propose to adopt SenticNet for ASA in a social media context. The assumption behind SenticNet is that semantically related concepts share common sentiment [67]. Therefore, the first step toward Arabic SenticNet is to extract Arabic concepts and to explore the semantic links between them. Extending literature studies with concept-based sentiment analysis is a promising direction. Works such as [117, 188] should be exploited in this direction. The authors in [117, 188] addressed frequently-used expressions to express sentiments such as prayers, idioms, and slangs. Furthermore, contests such as SemEval-2017 Task 4 (which focused on Twitter sentiment analysis [87, 177] and SemEval-2018 Task 1 (which focused on affect in tweets [8, 146]) can produce Arabic dialectical datasets that can aid the training of ASA models for social media analysis.

# 8 Conclusion and future work

In this paper, we reviewed the major works related to Arabic sentiment analysis. We found that sentiment analysis for Arabic has drawn increasing attention from the NLP research community. The two main approaches undertaken so far are monolingual and bilingual. While the monolingual method is based on Arabic sentiment resources to classify sentiment polarity, the bilingual method leverages English resources and machine translation. For the sentiment classification task, we noticed that three kinds of approaches are used: lexicon-based, corpus-based, and hybrid. The corpus-based and hybrid approaches are mainly constructed for machine learning algorithms. We found that, for Arabic, the most used algorithms are NB and SVM. Deep learning is not yet explored as much as for English sentiment analysis.

Experiments revealed that the performance strongly depends on the quality of sentiment resources. Existing Arabic sentiment resources are not good for social media analysis due to dialectical content. For other kinds of data, most of the studies reported that their findings are promising. However, many challenges need to be addressed to achieve effectiveness. We propose to further explore dialectical content available on social media to build comprehensive Arabic resources considering both Arabizi and dialects. Furthermore, we propose to shift from simple word-level sentiment analysis to concept-based sentiment analysis and to further explore word embedding to handle the complexity of Arabic. On the other hand, while analyzing works that have drawn the attention of researchers, we noticed that aspect-based sentiment analysis, opinion holder extraction, as well as spam detection and problems related to domain dependency, are not yet extensively explored. We hope that this survey can provide researchers with a comprehensive overview of Arabic sentiment analysis in terms of resources, approaches, and open challenges to further advance the state of the art in this domain.